\documentclass[sigconf]{acmart}

\usepackage{booktabs} 

\setcopyright{acmlicensed}
\usepackage{epstopdf}
\usepackage{caption}
\usepackage{multirow}
\usepackage[english]{babel}
\usepackage{amsmath}
\usepackage{algorithm}
\usepackage{algorithmic}
\usepackage{color}
\usepackage{url}

\begin{document}

\copyrightyear{2018} 
\acmYear{2018} 
\setcopyright{acmcopyright}
\acmPrice{15.00}
\acmDOI{10.1145/3159652.3159685}
\acmISBN{978-1-4503-5581-0/18/02}

\title{Modelling Domain Relationships for Transfer Learning on\\ Retrieval-based Question Answering Systems in E-commerce}


\author{Jianfei Yu}
\authornote{Work done during the internship at Alibaba Group.}
\affiliation{\institution{Singapore Management University}}
\email{jfyu.2014@phdis.smu.edu.sg}

\author{Minghui Qiu}
\authornote{Corresponding author.}
\affiliation{\institution{Alibaba Group, China}}
\email{minghui.qmh@alibaba-inc.com}

\author{Jing Jiang}
\affiliation{\institution{Singapore Management University}}
\email{jingjiang@smu.edu.sg}

\author{Jun Huang, Shuangyong Song}
\affiliation{\institution{Alibaba Group, China}}
\email{{junhuang.jh,shuangyong.ssy}@alibaba-inc.com}


\author{Wei Chu}
\affiliation{\institution{Alibaba Group, China}}
\email{weichu.cw@alibaba-inc.com}

\author{Haiqing Chen}
\affiliation{\institution{Alibaba Group, China}}
\email{haiqing.chenhq@alibaba-inc.com
}

\renewcommand{\shortauthors}{J. Yu, M. Qiu, J. Jiang, J. Huang, S. Song, W. Chu, and H. Chen.}

\begin{abstract}
Nowadays, it is a heated topic for many industries to build automatic question-answering (QA) systems.
A key solution to these QA systems is to retrieve from a QA knowledge base the most similar question of a given question, which can be reformulated as a paraphrase identification (PI) or a natural language inference (NLI) problem.
However, most existing models for PI and NLI have at least two problems: They rely on a large amount of labeled data, which is not always available in real scenarios, and they may not be efficient for industrial applications.

In this paper, we study transfer learning for the PI and NLI problems, aiming to propose a general framework, which can effectively and efficiently adapt the \textit{shared} knowledge learned from a resource-rich source domain to a resource-poor target domain. 
Specifically, since most existing transfer learning methods only focus on learning a shared feature space across domains while ignoring the relationship between the source and target domains, we propose to simultaneously learn shared representations and domain relationships in a unified framework.
Furthermore, we propose an efficient and effective hybrid model by combining a sentence encoding-based method and a sentence interaction-based method as our base model.
Extensive experiments on both paraphrase identification and natural language inference demonstrate that our base model is efficient and has promising performance compared to the competing models, and our transfer learning method can help to significantly boost the performance. 
Further analysis shows that the \textit{inter-domain} and \textit{intra-domain} relationship captured by our model are insightful.
Last but not least, we deploy our transfer learning model for PI into our online chatbot system, which can bring in significant improvements over our existing system.

Finally, we launch our new system on the chatbot platform Eva\footnote{\url{https://gcx.aliexpress.com/ae/evaenglish/portal.htm?pageId=195440}} in our E-commerce site \textit{AliExpress}\footnote{\url{http://www.aliexpress.com/}}.
\end{abstract}

%
%
%

\keywords{Retrieval-based Question Answering, Transfer Learning, Domain Relationships Learning, Adversarial Training}


\maketitle

\section{Introduction}
\label{intro}
Question Answering (QA) systems have been widely developed and used in many domains.
Examples of industry applications include Alibaba's AliME~\cite{qiu:acl17,alime-demo}, Microsoft's SuperAgent~\cite{cui:acl17}, Apple's Siri and Google's Google Assistant.
Generally speaking, there are two kinds of commonly-used techniques behind most QA systems: Information Retrieval (IR)-based models~\cite{yan:sigir2016} and generation-based models~\cite{vinyals:arxiv2015}.
In this work, we focus on building up an IR-based QA system for automatically answering frequently asked questions (FAQs) in the E-commerce industry.

\begin{figure}
\includegraphics[width=3.5in]{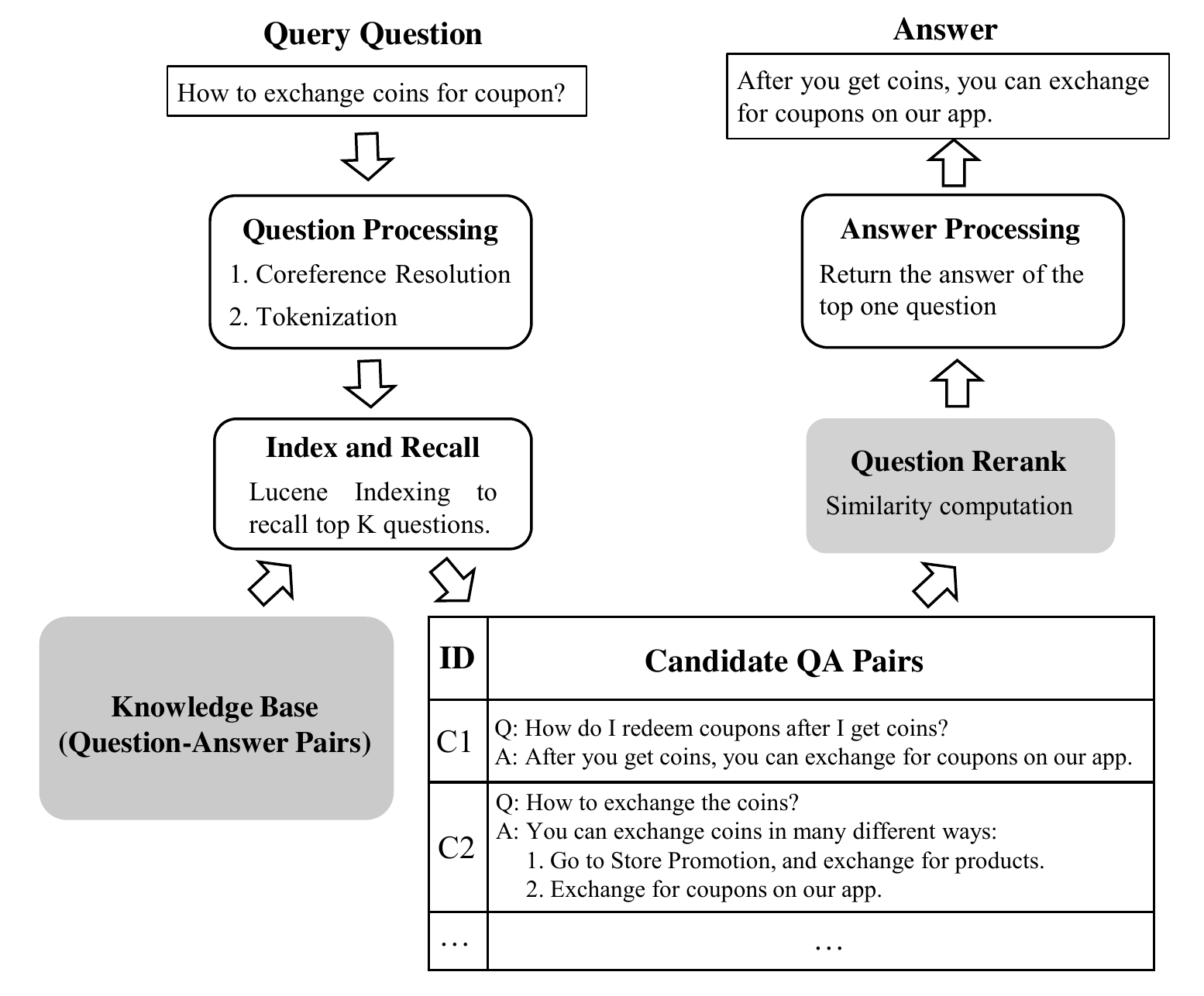}
\caption{The workflow of IR-based QA systems.}
\label{fig:examples}
\end{figure}

Fig.~\ref{fig:examples} illustrates the workflow of IR-based chatbot systems, where a key component is the \textit{Question Rerank} module which reranks candidate questions in a question-answering knowledge base (KB) to find the best matching question given a question from a user.
This task can be reduced to a paraphrase identification or a natural language inference problem.
Take the query question and knowledge base shown in Fig.~\ref{fig:examples} for example. 
If we can detect that question C1 in the KB is a paraphrase of the query question, then we can take its answer as the answer for the query.
In some cases, if we allow the matching question to be more general than the query question (i.e., entailed by the query question), we can also take the answer for question C2 in the KB as the query question's answer.

In the literature, paraphrase identification (PI) and natural language inference (NLI) have been extensively studied in the last decade~\cite{socher:nips2011,snli:emnlp2015,yin:tacl2016,bowman:acl2016,chen:acl2017}.
However, when applying existing solutions to PI and NLI in chatbot systems in the E-commerce industry, there are at least two major challenges we face:
(1) \textit{Lack of rich training data}: All these solutions rely on a large amount of labeled data. However, it is generally time-consuming and costly to manually annotate sufficient labeled data for each domain.
For example, different product categories might need different training data.
(2) \textit{Hard to reach a high QPS\footnote{Queries Per Second}}. Most of the existing methods focus on improving the effectiveness or accuracy without paying much attention to efficiency. 
For real industry applications, when real-time responses are expected and a large number of customers are being served simultaneously, we need an efficient method to support a high QPS.

\begin{figure}[!tp]
\includegraphics[height=4in, width=7in]{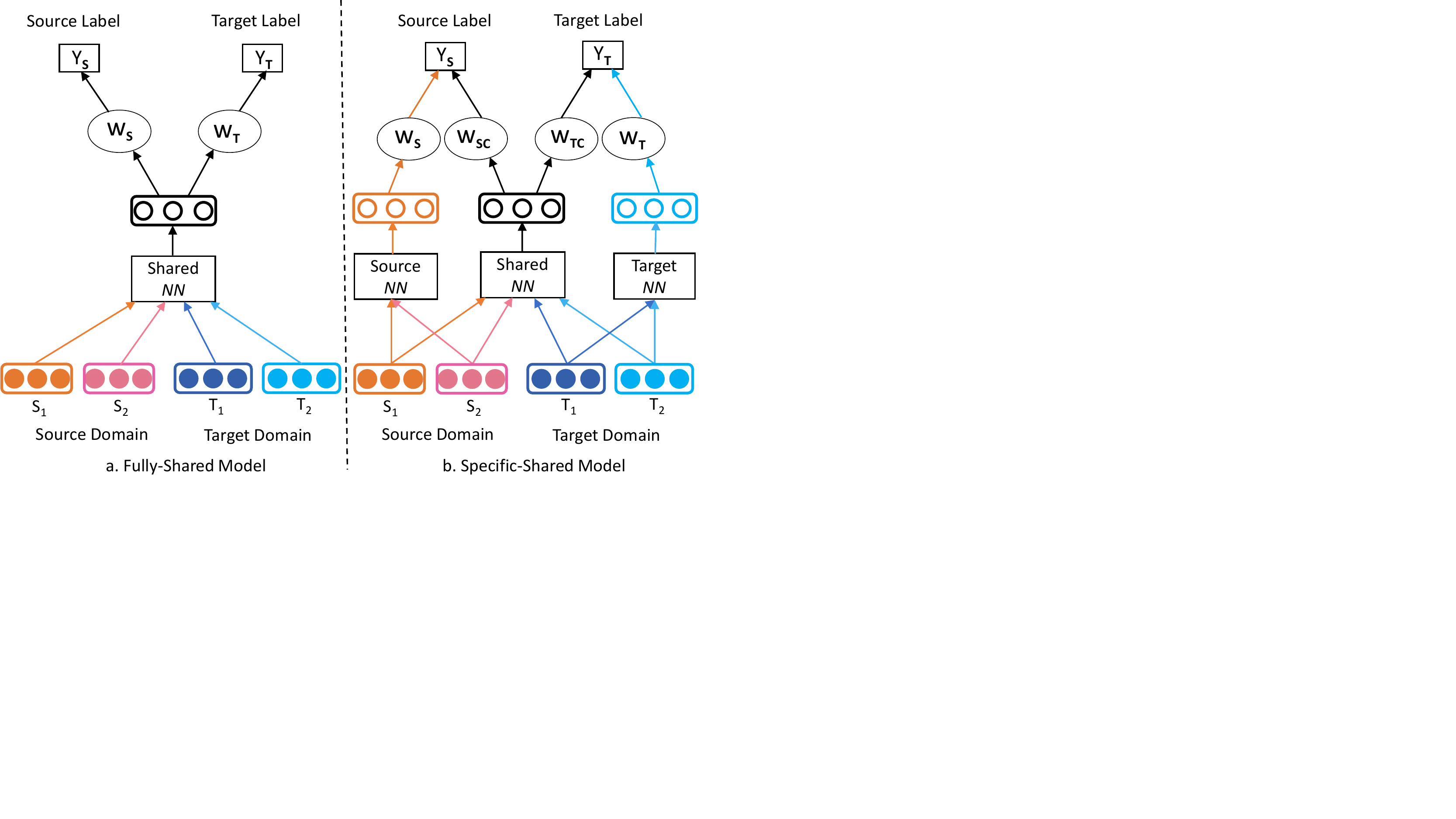}
\setlength{\abovecaptionskip}{-4.5cm}
\setlength{\belowcaptionskip}{-0.5cm}
\caption{Existing Transfer Learning Frameworks.}
\label{fig:tl}
\end{figure}

In this paper, we try to address the two challenges above.
Specifically, we first make an empirical comparison of both the effectiveness and efficiency of several representative methods for modeling sentence pairs and propose an effective and efficient \textit{hybrid model} as our base model.
This ensures that we can achieve a high QPS.
On top of the base model, we further design a new \textit{transfer learning (TL) framework}, which is able to efficiently improve the performance on a resource-poor target domain by leveraging knowledge from a resource-rich source domain.

\textbf{A Hybrid Base Model.} 
Observing that LSTM-based methods~\cite{snli:emnlp2015,chen:acl2017} are much more time-consuming than CNN-based methods~\cite{yin:tacl2016,mou:acl2016}, we focus on CNN-based methods in this study. Meanwhile, there are typically two types of CNN-based methods for the task, namely sentence encoding (SE)-based methods~\cite{yin:naacl2015,mou:EMNLP2016} and sentence interaction (SI)-based methods~\cite{hu:nips2014,pang:aaai2016}. We argue that these two types of methods may highly complement each other, and thus we propose a hybrid CNN model by combining an SE-based method~\cite{yin:tacl2016} and an SI-based method~\cite{pang:aaai2016}. Specifically, we modify the SE-based method using two element-wise comparison functions inspired by~\cite{mou:acl2016,wang:iclr2017} to match the two sentence embeddings, and then concatenate them together with sentence embeddings from the SI-based method.

\textbf{Transfer Learning Framework.} Transfer learning aims to apply knowledge gained in a source domain to help a target domain~\cite{pan:tkde2010}. The key issue is how to transfer the shared knowledge from the source domain to the target domain while exclude the specific knowledge in the source domain based on the \textit{domain relationship}.
Most recent studies for TL in NLP perform \textit{multi-task feature learning} by exploiting different NN models to capture a shared feature space across domains.
As illustrated in Fig.~\ref{fig:tl}a and Fig.~\ref{fig:tl}b, one line of work employs a fully-shared framework to learn a shared representation followed by using two different fully connected layers for each domain~\cite{mou:EMNLP2016,yang:iclr2017}, while another line of work uses a specific-shared framework to learn not only a shared representation for both domains but also a domain-specific representation for each domain~\cite{liu:acl2017}.



However, the first line of work simply assumes that two domains share the same feature space but ignore the domain-specific feature space.
Although the latter one is capable of capturing both the shared and the domain-specific representations, it does not consider any relationships between the weights of the final output layer.
Generally speaking, the weights on the output layer should capture both the \textit{inter-domain} and the \textit{intra-domain} relationships:
(1) For the shared feature space across domains, since it is expected to be domain-independent, the weights corresponding to this feature space in the two domains should be positively related to each other;
(2) For the shared and the domain-specific feature spaces in each domain, since they are expected to respectively capture domain-independent and domain-dependent features, their corresponding weights should be irrelevant to each other.
Motivated by such an intuition, in this paper, we propose a new transfer learning method by explicitly modeling the domain relationships via a covariance matrix, which imposes a regularization term on the weights of the output layer to uncover both the inter-domain and the intra-domain relationships.
Besides, to make the shared representation more invariant across domains, we follow some recent work on adversarial networks~\cite{ganin:jmlr2016,liu:acl2017} and introduce an adversarial loss on the shared feature space in our method.
Fig.~\ref{fig:base} gives an outline of our full model.



To evaluate our proposed method, we conduct both intrinsic evaluation and extrinsic evaluation.

\textbf{Intrinsic Evaluation.} We conduct extensive experiments on 
both a benchmark dataset and our own dataset. 
(1) The hybrid CNN model is shown to be not only efficient but also effective, in comparison with several representative methods;
(2) Our proposed transfer learning method can bring significant improvements over the base model without transfer learning, and outperform existing TL frameworks including the widely used fully-shared model and the recently proposed specific-shared model;
(3) Further analysis on our learned correlation matrix shows that our method is able to capture the inter-domain and intra-domain relationships.

\textbf{Extrinsic Evaluation.}
We deploy our proposed hybrid CNN-based transfer learning model into our online chatbot system, which is deployed on a real E-commerce site \textit{AliExpress}.
Both the offline and the online evaluations show that our new system can significantly outperform the existing online chatbot system.
Finally, we launch our new system on Eva\footnote{Eva can be accessed via the following link: \url{https://gcx.aliexpress.com/ae/evaenglish/portal.htm?pageId=195440}}, a chatbot platform in \textit{AliExpress}.

\section{Model}
\label{body}


In this section, we present our general model for paraphrase identification and natural language inference, which will be used for question reranking in our chatbot-based QA system.

\subsection{Problem Formulation and Notation}
\label{base-model}
Our model is designed to address the following general problem. 
Given a pair of sentences, we would like to identify their semantic relation. 
For paraphrase identification (PI), the semantic relation indicates whether or not the two sentences express the same meaning~\cite{yin:naacl2015}; for natural language inference (NLI), it indicates whether a hypothesis sentence can be inferred from a premise sentence~\cite{snli:emnlp2015}.

Formally, assume there are two sentences $\mathbf{X}_1 = (\mathbf{x}^{1}_{1}, \mathbf{x}^{1}_{2}, \ldots, \mathbf{x}^{1}_{m})$ and $\mathbf{X}_2 = (\mathbf{x}^{2}_{1}, \mathbf{x}^{2}_{2}, \ldots, \mathbf{x}^{2}_{n})$, where $\mathbf{x}_{i}^j$ denotes an $l$-dimensional dense embedding vector retrieved from a lookup table $\mathbf{E} \in \mathbb{R}^{l \times |\mathcal{V}|}$ for all the words in the vocabulary $\mathcal{V}$.
Our task is to predict the semantic label $y$ which indicates the relation between $\mathbf{X}_1$ and $\mathbf{X}_2$.
For PI, we assume the label $y$ to be either \textit{paraphrase} or \textit{not paraphrase}; for NLI, we assume $y$ to be either \textit{neutral}, \textit{entailment} or \textit{contradiction}.

We consider a transfer learning setting, where we have a set of labeled sentence pairs from a source domain and a target domain, respectively, denoted by $\mathcal{D}^\text{s}$ and $\mathcal{D}^\text{t}$.
Note that $|\mathcal{D}^\text{s}|$ is assumed to be much larger than $|\mathcal{D}^\text{t}|$.
We seek to use both $\mathcal{D}^\text{s}$ and $\mathcal{D}^\text{t}$ to train a good model so that it can work well in the target domain.

To solve such a problem, a widely used transfer learning method~(as illustrated in Fig.~\ref{fig:tl}a) is to use the same NN model to transform every pair of input sentences in both domains into a hidden representation $\mathbf{z_c} \in \mathbb{R}^{q}$, where $q$ is the size of the hidden representations.
To facilitate our discussion, let us assume $\mathbf{z_c} = \textit{f}_{\Theta_c}(\mathbf{X}_1, \mathbf{X}_2)$, where $\textit{f}_{\Theta_c}$ denotes the transformation function parameterized by $\Theta_c$.
Next, for the source and the target domains, we assume that two fully connected layers are separately learned to map $\mathbf{z_c}$ to label $y$. 
\[ p(y \mid \mathbf{z_c}) =
  \begin{cases}
    \text{softmax}(\mathbf{W_{sc}} \mathbf{z_c}+\mathbf{b_s})      & \quad \text{if } y \text{ from src,}\\
    \text{softmax}(\mathbf{W_{tc}} \mathbf{z_c}+\mathbf{b_t})      & \quad \text{if } y \text{ from tgt,}\\
  \end{cases}
\]
where $\mathbf{W_{sc}} \in \mathbb{R}^{|Y| \times q}$ and $\mathbf{W_{tc}} \in \mathbb{R}^{|Y| \times q}$ are weight matrices and $\mathbf{b_s} \in \mathbb{R}^{|Y|}$ and $\mathbf{b_t} \in \mathbb{R}^{|Y|}$ are bias vectors.

Besides, another transfer learning approach~\cite{liu:acl2017} was recently proposed to use a domain-shared NN model and two domain-specific NN models to obtain a shared embedding $\mathbf{z_c}$ and two domain-specific embeddings $\mathbf{z_s}= \textit{f}_{\Theta_s}(\mathbf{X}_1, \mathbf{X}_2)$ and $\mathbf{z_t}= \textit{f}_{\Theta_t}(\mathbf{X}_1, \mathbf{X}_2)$.
Therefore, the output layers are defined as: 
\[ p(y \mid \mathbf{z_c}, \mathbf{z_s}) =
  \begin{cases}
    \text{softmax}(\mathbf{W_{sc}} \mathbf{z_c}+ \mathbf{W_s} \mathbf{z_s} + \mathbf{b_s})
    & $ $ \text{if } y \text{ from src,}\\
    \text{softmax}(\mathbf{W_{tc}} \mathbf{z_c}+ \mathbf{W_t} \mathbf{z_t} +\mathbf{b_t})
    & $ $ \text{if } y \text{ from tgt.}\\
  \end{cases}
\]

The main limitation of the fully-shared framework is that it ignores source-specific or target-specific features. 
While for the specific-shared framework, it fails to consider any inherent correlations between the weights on the output layers.
Therefore, we will introduce our proposed method that explicitly incorporates such correlations into the specific-shared framework in the next session.

\subsection{Proposed Transfer Learning Method}
\label{tl-model}

Our goal is to model the inter-domain relationship between $\mathbf{W_{sc}}$ and $\mathbf{W_{tc}}$, and the intra-domain relationship between $\mathbf{W_{s}}$ and $\mathbf{W_{sc}}$ as well as $\mathbf{W_{t}}$ and $\mathbf{W_{tc}}$.
Hence, we first reshape each weight matrix into a vector $\mathbf{w_{d}} \in \mathbb{R}^{q|Y|}$, followed by concatenating all the four reshaped vectors to form a new matrix $\mathbf{W} \in \mathbb{R}^{q|Y| \times 4}$, where each column corresponds to one weight matrix of the output layer.

Next, to capture the domain relationships mentioned above, we introduce a covariance matrix $\mathbf{\Omega} \in \mathbb{R}^{4 \times 4}$.
Note that each element $\mathbf{\Omega_{i,j}}$ indicates the correlation between $\mathbf{W_{i}}$ and $\mathbf{W_{j}}$, where $\mathbf{i}$ and $\mathbf{j}$ are one of $\mathbf{s}$, $\mathbf{sc}$, $\mathbf{t}$ and $\mathbf{tc}$.
Inspired by a general multi-task relationship learning framework as introduced in~\cite{zhang:uai2010}, we consider confining the output layer's weights with $\mathbf{\Omega}$ by using $\textbf{tr}(\mathbf{W}\mathbf{\Omega}^{-1}\mathbf{W}^T)$, where $\textbf{tr}(\cdot)$ is the trace of a square matrix.
This means that if $\mathbf{\Omega_{i,j}}$ is a large positive/negative value, $\mathbf{W_{i}}$ and $\mathbf{W_{j}}$ will be positively/negatively related to each other; otherwise if $\mathbf{\Omega_{i,j}}$ is close to zero, $\mathbf{W_{i}}$ and $\mathbf{W_{j}}$ will be irrelevant to each other.
Note that to the best of our knowledge, we are the first to apply the multi-task relationship learning framework into NN-based transfer learning methods.

In order to simultaneously learn our model parameters and the domain relationships in a unified framework, we formulate our loss function as follows:
\begin{align}
\label{eqn:loss1}
\notag\mathcal{L} =  & \sum_{\textbf{k} \in {\mathbf{s},\mathbf{t}}} - \frac{1}{n_{\textbf{k}}}\sum^{n_{\textbf{k}}}_{i=1} \log p\big(y_{i} \mid \mathbf{X}^i_1, \mathbf{X}^i_2 \big) + \frac{\lambda_1}{2} \textbf{tr}(\mathbf{W}\mathbf{\Omega}^{-1}\mathbf{W}^T)\\
\notag& + \frac{\lambda_2}{2} ||\mathbf{W}||^2_F + \frac{\lambda_3}{2} ||\Theta_c||^2_F + \frac{\lambda_4}{2} ||\Theta_s||^2_F + \frac{\lambda_5}{2} ||\Theta_t||^2_F\\
\text{s.t.} & \quad \mathbf{\Omega} \geq 0, \quad \textbf{tr}(\mathbf{\Omega}) = 1.
\end{align}
where $\lambda_1$, $\lambda_2$, $\lambda_3$, $\lambda_4$ and $\lambda_5$ are regularization parameters, $\mathbf{\Omega}$ is required to be positive semi-definite, and $\textbf{tr}(\mathbf{\Omega})$ is required to be 1 without losing generality.
In the above formulation, the first term refers to the cross-entropy loss for both domains, and the second term serves as a domain-relationship regularizer to constrain the weights on the output layer. The remaining terms are standard L2-regularization terms.

\subsection{Adversarial Loss}
\label{adv-model}
Our above transfer learning method is based on the specific-shared framework, which is assumed to well capture the shared and domain-specific feature spaces.
However, as suggested by~\cite{liu:acl2017}, the shared representation learned in this framework may still contain noisy domain-specific features.
Therefore, to eliminate the noisy features, here we also consider incorporating an adversarial loss on the shared feature space so that the trained model can not distinguish between the source and target domains on it~\cite{ganin:jmlr2016}.

First, we assume that the shared layer $\mathbf{z_c}$ is mapped to a binary domain label $d$, which indicates whether $\mathbf{z_c}$ comes from the source or the target domain:
\begin{eqnarray*}
p(d \mid \mathbf{z_c}) = \text{softmax}(\mathbf{W_d} \mathbf{z_c}+\mathbf{b_d}).
\end{eqnarray*}
Since the goal of adversarial training is to encourage the shared feature space indiscriminate across two domains, we define the adversarial loss as minimizing the negative entropy of the predicted domain distribution, which is different from maximizing the negative cross-entropy as in~\cite{ganin:jmlr2016,liu:acl2017}:
\begin{align}
\label{eqn:advloss}
\boldsymbol\ell = & \sum_{\textbf{k} \in {\mathbf{s},\mathbf{t}}} \Big(\frac{1}{n_\textbf{k}}\sum^{n_\textbf{k}}_{i=1}\sum^1_{j=0} p(d_{ij} \mid \mathbf{X}^i_1, \mathbf{X}^i_2) \log p(d_{ij} \mid \mathbf{X}^i_1, \mathbf{X}^i_2)\Big).
\end{align}

Finally, we obtain a combined objective function as follows:
\begin{align*}
&\quad \min_{\substack{ \mathbf{\Omega},\mathbf{W},\mathbf{W_d}, \Theta_{c}, \\ \Theta_{s}, \Theta_{t}, \mathbf{b_s},\mathbf{b_t}, \mathbf{b_d}}} \mathcal{L} + \lambda_0 \boldsymbol\ell \\
\notag\text{s.t.} & \qquad\quad \mathbf{\Omega} \geq 0, \quad \textbf{tr}(\mathbf{\Omega}) = 1,
\end{align*}
where $\lambda_0$ is a hyper-parameter for tuning the importance of the adversarial loss.
As suggested by~\cite{zhang:uai2010}, it is not easy to optimize such a semi-definite programming problem. 
We will present an alternating training approach in Section~\ref{infer} for solving it efficiently.

\subsection{Base Model}
\label{basic-model}

Although the proposed transfer learning method is general and any neural networks for modeling a pair of sentences can be applied to it, we further target at proposing an efficient and effective base model for encoding a pair of sentences.

On one hand, although various attention-based LSTM architectures have been proposed to achieve a superior performance on both PI and NLI~\cite{rock:iclr2015,parikh:emnlp2016,wang:iclr2017,chen:acl2017}, these models are very time-consuming due to the computation of memory cells and attention weights in each time step, which may not satisfy the industry demand, especially when QPS is high.
On the other hand, CNN-based models are proven to be efficient, hence are the focus of our study. Most existing CNN-based models can be categorized into two groups: sentence encoding (SE)-based methods and sentence interaction (SI)-based methods.
The former aims to first learn good representations for each sentence, followed by using a comparison function to transform them into a single representation~\cite{yin:naacl2015,mou:EMNLP2016}, while the latter tries to directly model the interaction between two sentences at the beginning and then makes abstractions on top of the interaction output~\cite{hu:nips2014,pang:aaai2016}.
Observing that the two lines of methods focus on different perspectives to model sentence pairs, we expect that a combination of them can capture both good sentence representations and rich interaction structures.

\begin{figure*}
\includegraphics[height=4in, width=7in]{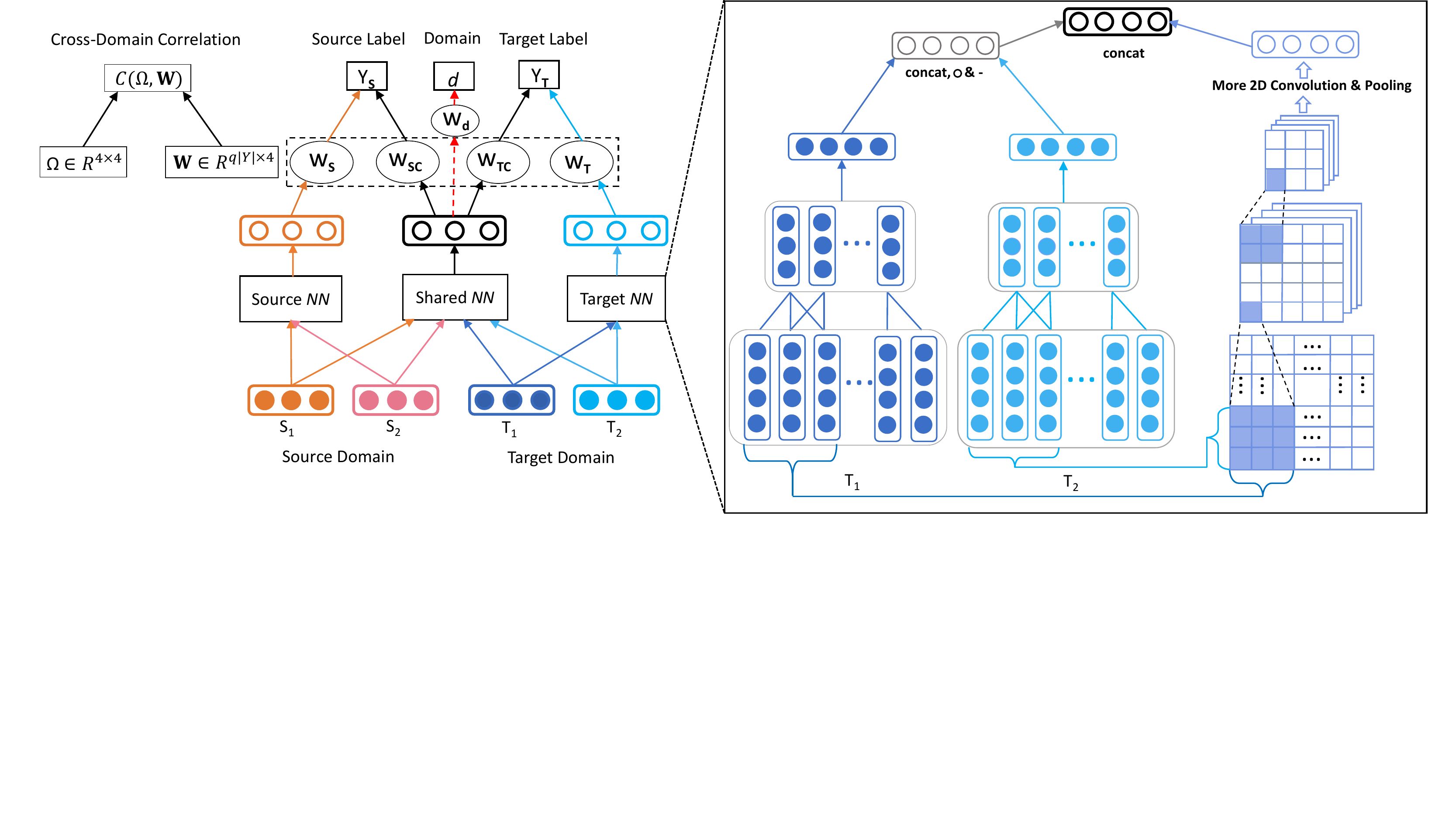}
\setlength{\abovecaptionskip}{-3.9cm}
\setlength{\belowcaptionskip}{-0.2cm}
\caption{Our Full Transfer Learning Model for Paraphrase Identification and Natural Language Inference.}
\label{fig:base}
\end{figure*}

Hence, we propose a hybrid CNN (hCNN) model, which are based on some minor modifications of two existing models: a SE-based BCNN model~\cite{yin:tacl2016} and a SI-based Pyramid model~\cite{pang:aaai2016}.
Fig.~\ref{fig:base} depicts our full transfer learning framework, which contains one shared hCNN and two domain-specific hCNNs.
Below we briefly go through the architecture of hCNN.
Note that in our implementation and the model description below, we pad the two input sentences to the same length $m$.

\textbf{Modified BCNN:}
Following the original BCNN~\cite{yin:tacl2016}, we first use two separate 1-D convolutional (conv) and 1-D max-pooling layers to encode the two input sentences into two sentence embeddings:
\begin{eqnarray*}
\mathbf{h_1} = \textit{CNN}(\mathbf{X}_{1}); \quad \mathbf{h_2} = \textit{CNN}(\mathbf{X}_{2}).
\end{eqnarray*}
Furthermore, as suggested by~\cite{mou:acl2016,wang:iclr2017} that element-wise comparison can work well on the problem, we use two comparison functions to match the two sentence embeddings, and then concatenate them together with the sentence embeddings as the sentence pair representation:
$\mathbf{H_b} = \mathbf{h_1} \oplus \mathbf{h_2} \oplus (\mathbf{h_1}-\mathbf{h_2}) \oplus (\mathbf{h_1} \cdot \mathbf{h_2}),
$
where $-$ and $\cdot$ refer to element-wise subtraction and element-wise multiplication, and $\oplus$ refers to concatenation.
Note that this setting is different from the original BCNN, which yields better performance in our empirical experiments.

\textbf{Pyramid:}
As shown in the rightmost part of Fig \ref{fig:base}, we first produce an interaction matrix $\mathbf{M} \in \mathbb{R}^ {m \times m}$, where $\mathbf{M_{i,j}}$ denotes the similarity score between the $i^\text{th}$ word in $\mathbf{X}_1$ and the $j^\text{th}$ word in $\mathbf{X}_2$.
Following~\cite{pang:aaai2016}, we use dot-product to compute the similarity score. 

Next, by viewing the interaction matrix as an image, we stack two 2-D convolutional layers and two 2-D max-pooling layers on it to obtain the hidden representation $\mathbf{H_p}$.

Finally, we concatenate the two hidden representations as the final representation for each input sentence pair: $\mathbf{z} = \mathbf{H_b} \oplus \mathbf{H_p}$.

\subsection{Inference}
\label{infer}

In our combined objective function, we have nine parameters $\mathbf{\Omega}$, $\mathbf{W}$, $\mathbf{W_d}$, $\mathbf{b_s}$, $\mathbf{b_t}$, $\mathbf{b_d}$, $\Theta_{c}$, $\Theta_{s}$ and $\Theta_{t}$, and it is not easy to optimize them at the same time.
Following the practice in~\cite{zhang:uai2010}, we employ an alternating stochastic method, i.e., first optimizing the other eight parameters by fixing $\mathbf{\Omega}$, and then alternatively optimizing $\mathbf{\Omega}$ by fixing the others in each iteration.
The details are given as below:

\textbf{Updating $\mathbf{W}$, $\mathbf{W_d}$, $\mathbf{b_s}$, $\mathbf{b_t}$, $\mathbf{b_d}$, $\Theta_{c}$, $\Theta_{s}$ and $\Theta_{t}$}.
While fixing $\mathbf{\Omega}$, the optimization problem becomes:
\begin{align}
\label{eqn:loss1}
\notag& \min_{\substack{\mathbf{W},\mathbf{W_d},\Theta_{c},\Theta_{s},\\ \Theta_{t}, \mathbf{b_s},\mathbf{b_t},\mathbf{b_d}}} \sum_{\textbf{k} \in {\mathbf{s},\mathbf{t}}}  \Big(\frac{1}{n_\textbf{k}}\sum^{n_\textbf{k}}_{i=1} \big(-\log p(y_{i} \mid \mathbf{X}^i_1, \mathbf{X}^i_2)\\
\notag& + \lambda_0 \sum^1_{j=0} p(d_{ij} \mid \mathbf{X}^i_1, \mathbf{X}^i_2) \log p(d_{ij} \mid \mathbf{X}^i_1, \mathbf{X}^i_2) \big)\Big)+ \frac{\lambda_1}{2} \textbf{tr}(\mathbf{W}\mathbf{\Omega}^{-1}\mathbf{W}^T) \\
\notag& + \frac{\lambda_2}{2} ||\mathbf{W}||^2_F + \frac{\lambda_3}{2} ||\Theta_c||^2_F + \frac{\lambda_4}{2} ||\Theta_s||^2_F + \frac{\lambda_5}{2} ||\Theta_t||^2_F
\end{align}
Since it is a smooth function, we can easily compute its partial derivatives with respect to the eight parameters.

\textbf{Updating $\mathbf{\Omega}$}.
After fixing the eight parameters, the optimization problem is as follows:
\begin{align}
\notag\min_{\mathbf{\Omega}} & \quad  \textbf{tr}(\mathbf{W}\mathbf{\Omega}^{-1}\mathbf{W}^T)\\
\notag\text{s.t.} & \quad \mathbf{\Omega} \geq 0, \quad \textbf{tr}(\mathbf{\Omega}) = 1.
\end{align}
As proved by~\cite{zhang:uai2010}, the above optimization problem has an analytical solution $\mathbf{\Omega} = \frac{(\mathbf{W}^T\mathbf{W})^{\frac{1}{2}}}{\textbf{tr}\big((\mathbf{W}^T\mathbf{W})^{\frac{1}{2}}\big)}$.

Finally, we present the whole procedure for training our full model as in Algorithm \ref{alg:infer}.
Note that we only update $\mathbf{\Omega}$ when we scan all the target training instances once.

\begin{algorithm}[!tp]                   
\caption{Training Procedure for our Full Model}          
\label{alg:infer}                           
\begin{algorithmic}[1]                    
    \STATE \textbf{Input:} source training instances $\mathcal{D}^\text{s}$, target training instances $\mathcal{D}^\text{t}$ ($|\mathcal{D}^\text{s}| \gg |\mathcal{D}^\text{t}|$) .
    \STATE \textbf{Output:} $\mathbf{\Omega}$, $\mathbf{W}$, $\mathbf{W_d}$, $\mathbf{b_s}$, $\mathbf{b_t}$, $\mathbf{b_d}$, $\Theta_{c}$, $\Theta_{s}$, $\Theta_{t}$
    \STATE Initialize $\mathbf{W}$, $\mathbf{W_d}$, $\mathbf{b_s}$, $\mathbf{b_t}$, $\mathbf{b_d}$, $\Theta_{c}$, $\Theta_{s}$, $\Theta_{t}$ with random values
    \STATE Initialize $\mathbf{\Omega} = \frac{1}{4}\mathbf{I_4}$, where $\mathbf{I}$ is an identity matrix
    \STATE $\text{epoch} = 0$
    \WHILE{$\text{epoch} \leq \text{MaxEpoch}$}
        \STATE {$c_s$, $c_t$ = 0, 0}
        \WHILE{$c_s$ $\leq$ src\_batches}
            \STATE {read the ${c_s}^\text{th}$ mini-batch from the source domain}
            \STATE {Update $\Theta_{c}$, $\Theta_{s}$, $\mathbf{W}$, $\mathbf{W_d}$, $\mathbf{b_s}$ and $\mathbf{b_d}$}
            \STATE {$c_s$ += 1}
            \IF {$c_t$ == tgt\_batches}
                \STATE {$c_t$ = 0}
                \STATE {Update $\mathbf{\Omega} = \frac{(\mathbf{W}^T\mathbf{W})^{\frac{1}{2}}}{\textbf{tr}\big((\mathbf{W}^T\mathbf{W})^{\frac{1}{2}}\big)}$}
            \ENDIF
            \STATE {read the ${c_t}^\text{th}$ mini-batch from the target domain}
            \STATE {Update $\Theta_{c}$, $\Theta_{t}$, $\mathbf{W}$, $\mathbf{W_d}$, $\mathbf{b_t}$ and $\mathbf{b_d}$}
            \STATE {$c_t$ += 1}
        \ENDWHILE
        \STATE {epoch = epoch + 1}
    \ENDWHILE
\end{algorithmic}
\end{algorithm}

\subsection{Implementation Details}
\label{details}

In our full transfer learning model, we initialize the lookup table $\mathbf{E}$ with the pre-trained vectors from \textit{GloVe}~\cite{pennington:emnlp2014} by setting $l$ as 300. 
For \textbf{BCNN}, the window size and activation function are set to be 4 and ReLU, and the feature map sizes are set to be 50 and 100 for PI and NLI;
for the two convolution layers of \textbf{Pyramid} in both PI and NLI, the feature map sizes are set to be 8 and 16, the strides are set to be 1 and 3, and the kernel sizes are set to be $6\times 6$ and $4\times 4$; for the two max-pooling layers of \textbf{Pyramid}, the strides are set to be 4 and 2, and the pooling sizes are set to be $4\times 4$ and $2\times 2$.
Besides, for $\lambda_0$ and $\lambda_1$, we set them as 0.05 and 0.0008; while for $\lambda_2$, $\lambda_3$, $\lambda_4$ and $\lambda_5$, we set them as 0.0004.
AdaGrad~\cite{duchi:jmlr2011} is used to train our model with an initial learning rate of 0.08.

\vspace*{-.1cm}
\section{Online System}
\label{deploy}

As introduced in Section~\ref{intro}, our online chatbot system is based on traditional information retrieval techniques, where the goal is to obtain the nearest question in the knowledge base for a given customer question~\cite{jeon:cikm2005}.
Fig.~\ref{fig:examples} depicts the whole system architecture.

Specifically, we first build an indexing for all the questions in our knowledge base (KB) using \textit{Apache Lucene}~\footnote{\url{https://lucene.apache.org/core/}}.
Next, given a query question, we employ TF-IDF ranking algorithm~\cite{wu:tois2008} in Lucene to compute its similarities to all the questions in the KB, and call back the top-$K$ candidate questions.
We then use a reranking algorithm to compute the similarities between the query and the $K$ candidates, and  obtain the most similar candidate. 
Finally we return the answer of the selected candidate to answer the query question.
Note that in this paper, we only consider formulating our question rerank module as a PI task, but one can also model it as an NLI task.

Our existing reranking method is based on this ensemble method for the Answer Selection task~\cite{wang:acl2015}.
But instead of using the output of the time-consuming LSTM model, we feed another three features, namely, Word Mover's Distance~\cite{kusner:icml2015}, keywords features~\cite{wang:acl2015} and the cosine distance of sentence embeddings~\cite{wieting:iclr2016} to a gradient boosted regression tree (GBDT).

To combine our model with the existing ranking method, we treat the probability of being \textit{paraphrases} predicted by our model as an additional feature, and feed all features to GBDT for reranking.

\section{Experiments}
\label{exp}

In this section, we describe a qualitative evaluation of our proposed methods from the following perspectives:
(1) From Section~\ref{exp-setting} to Section~\ref{domrel}, we perform an \textbf{intrinsic} evaluation by utilizing a benchmark dataset and our own dataset to show the efficiency and effectiveness of our proposed base model and transfer learning framework;
(2) In Section~\ref{onlineeva}, we deploy our full model 
into our chatbot system, and conduct an \textbf{extrinsic} evaluation to show that our full model can bring in significant improvements to our existing online chatbots. 

\subsection{Experiment Settings}
\label{exp-setting}

\begin{table*}[!tp]
\setlength{\tabcolsep}{3.0pt}
\begin{minipage}[t]{.51\linewidth}
\centering
\medskip
{\footnotesize
\setlength{\abovecaptionskip}{0.05cm}
\caption{\small Statistics of Paraphrase Identification Data}
\label{tab:qq-stat}
\begin{tabular}{@{}lllll@{}}
\toprule
                            &           & Train        & Dev & Test      \\\midrule
                            & Q-Q Pairs & 29,884/8,624  & 7,622/1,968  & 7,569/2,133 \\
\multirow{3}{*}{AliExpress} & \#Query-Q & 3202/2414   & 781/578  & 777/584   \\
                            & \#Candi-Q & 9.33         & 9.76  & 9.74      \\
                            & \#words /Query-Q & 11.71         & 11.73  & 12.04      \\
                            & \#words /Candi-Q & 8.37         & 8.37  & 8.54      \\
                            \midrule
Quora                       & Q-Q Pairs & 404,290/149,265 & N.A.  & N.A.     \\\bottomrule
\end{tabular}}
\linespread{1.5}
\end{minipage}
\begin{minipage}[t]{0.48\linewidth}
\centering
\medskip
\setlength{\abovecaptionskip}{0.05cm}
\caption{\small Statistics of Natural Language Inference Data}
\label{tab:snli-stat}
\begin{tabular}{@{}lllllll@{}}
\toprule
      & SNLI   & Fiction & Travel & Slate      & Telephone & Government \\ [0.6ex] \midrule
Train & 550,125 & 77,348   & 77,350  & 77,306      & 83,348     & 77,350 \\ [1.2ex]
Dev   & 10,000  & 2,000    & 2,000   & 2,000       & 2,000      & 2,000  \\ [1.2ex]
Test  & 10,000  & 2,000    & 2,000   & 2,000       & 2,000      & 2,000  \\ [1.2ex]
\bottomrule
\end{tabular}
\linespread{0.9}
\end{minipage}
\end{table*}

\noindent  \textbf{Datasets:}
In this section, we evaluate our methods on both Paraphrase Identification (PI) and Natural Language Inference (NLI).

For PI, we used a recently released large-scale dateset\footnote{https://www.kaggle.com/c/quora-question-pairs} by Quora as the source domain, and our E-commerce dataset as the target domain.
Based on our historical data, we constructed a question answering KB, which consists of around 15,000 frequently asked QA pairs.
To create labeled question pairs, we first collected all the query questions from the chat log of conversations between clients and our staff from May 22 to May 28, 2017.
For each query question, we then used \textit{Lucene} indexing to retrieve several of its similar questions, and obtained 45,075 question pairs.
Finally, we asked a business analyst to annotate all the question pairs.

For NLI, we employed a large-scale multi-genre corpus~\cite{williams:arxiv2017}, which contains an image captioning domain (SNLI) and
another five distinct genres/domains about written and spoken English (MultiNLI)\footnote{For the MultiNLI dataset, we use Version 0.9 in this paper. Note that since the label of the original test set is unavailable, we treat its development set as our test set, and randomly choose 2000 sentence pairs from its training set as our development set.}.
Since the number of sentence pairs in SNLI is much larger than that in the other five domains, we took SNLI as the source domain, and the others as the target domains.

Table~\ref{tab:qq-stat} and Table~\ref{tab:snli-stat} summarize the statistics of our datasets.
Note that in Table~\ref{tab:qq-stat}, the number before and after the slash for Q-Q pairs denote respectively the total number of question pairs and the number of positive question pairs (i.e., paraphrases), while the two numbers for \#Query-Q respectively denote the total number of query questions and the number of questions with paraphrasing candidates. Besides, for \#Candi-Q, we refer to the average number of candidate questions for each query.

\noindent  \textbf{Compared Methods:}
For base models, we compared our hCNN model with the following models:
\begin{itemize}
\item \textbf{BCNN} is the left component of our hCNN model, which incorporates element-wise comparisons on top of the base model proposed in ~\cite{yin:tacl2016}.
\item \textbf{Pyramid} is the right component of our hCNN model based on sentence interactions as in~\cite{pang:aaai2016}.
\item \textbf{ABCNN} is the attention-based CNN model by~\cite{yin:tacl2016}.
\item \textbf{BiLSTM} is similar to \textbf{BCNN}, but uses LSTM instead of CNN to encode each sentence as in~\cite{snli:emnlp2015}.
\item \textbf{ESIM} is one of the state-of-the-art attention-based LSTM models on SNLI proposed by~\cite{chen:acl2017}.
\item \textbf{hCNN} is our hybrid CNN model as introduced in Section~\ref{basic-model}.
\end{itemize}

For evaluating the proposed transfer learning framework, we employed the following compared systems:
\begin{itemize}
\item \textbf{Tgt-Only} is the baseline trained in the target domain. 
\item \textbf{Src-Only} is another baseline trained in the source domain. 
\item \textbf{Mixed} is to simply combine the labeled data in the two domains to train the hCNN model.
\item \textbf{Fine-Tune} is a widely used TL method, where we first train a model on the source data, and then use the learned parameters to initialize the model parameters for training another model on the target data.
\item \textbf{FS} and \textbf{SS} are the fully-shared and specific-shared frameworks as detailed in Section~\ref{base-model}.
\item \textbf{DRSS} is our proposed model of learning domain relationships based on \textbf{SS} as in Section~\ref{tl-model}.
\item \textbf{SS-Adv} and \textbf{DRSS-Adv} denote adding the adversarial loss into \textbf{SS} and \textbf{DRSS} as in Section~\ref{adv-model}.
\end{itemize}

All the methods in this paper are implemented with Tensorflow and are trained on machines with NVIDIA Tesla K40m GPU.

\noindent  \textbf{Evaluation Metrics:}
For PI, since our goal is to retrieve the most similar candidate for each query question, we use our model to predict each candidate's probability of being \textit{paraphrase} as its similarity score, and then rank all the candidates.
To evaluate the ranking performance, we use Precision@1, Recall@1, $\text{F}_1$@1 as metrics;
to evaluate the classification performance for all question pairs, we employ two metrics: the Area under the Receiver Operating Characteristic curve (AUC) score~\cite{bradley:pr1997} and the classification accuracy (ACC).
For NLI, we only use ACC as the evaluation metric.

\subsection{Comparisons Between Base Models}
\begin{table}[h!]
\centering
\setlength{\abovecaptionskip}{0.05cm}
\caption{\small A Comparison Between Different Base Models}
\label{tab:indomain}
\small
\begin{tabular}{@{}llllll@{}}
\toprule
        & \multicolumn{3}{c}{E-Commerce} & \multicolumn{2}{c}{SNLI} \\
        \cmidrule(r){2-4} \cmidrule(r){5-6}
               & AUC   & ACC   & Test Time(ms)  & ACC      & Test Time(ms)     \\ \midrule
BCNN           & 81.0  & 77.5  & 2.8            & 81.0     & 3.3               \\
Pyramid        & 77.8  & 77.0  & 3.8            & 77.7     & 5.3               \\
ABCNN          & 81.0  & 78.2  & 5.2            & 81.8     & 12.3              \\
\textbf{hCNN}  & 82.2$\dagger$   & 79.2$\dagger$  & 4.3            & 83.2$\dagger$     & 6.4               \\ \midrule
BiLSTM         & 79.9  & 77.8  & 7.1            & 80.6     & 19.6              \\
ESIM           & 84.2  & 79.8  & 32.2           & 86.7     & 79.5              \\\bottomrule
\end{tabular}
\end{table}
In Table~\ref{tab:indomain}, we compared different models for classifying sentence pairs with hCNN in both efficiency and effectiveness.
Note that to fairly evaluate the efficiency of each model, we compute the total time of predicting all the test sentence pairs on CPU by setting the mini-batch size to 1, and report the average time.
Also, for feature map sizes in BCNN, ABCNN and the BCNN component in hCNN, we set them as 50 for PI and 300 for NLI.

First, we can find that LSTM-based methods are generally much slower than CNN-based methods.
Especially for ESIM, although it can outperform all CNN-based models, its computational time for each sentence pair is 32.2ms for our dataset and 79.5ms for SNLI, which is 6-11 times of CNN-based models.
This means that most existing state-of-the-art models can only support low QPS, and therefore hard to be applied to industry.
Second, clearly for both tasks, hCNN performs better than the other CNN-based methods, which indicates that BCNN and Pyramid are complementary to each other, and can work better when combined.
Moreover, we verified that the improvements of hCNN over the other methods are significant with $p < 0.05$ based on McNemar's paired significance test~\cite{gillick:icassp1989}.
Finally, while the computational cost of hCNN is slightly higher than BCNN and Pyramid, it can still serve 233 question pairs per second, which is able to satisfy the current demand of our industrial bot.

\subsection{Comparisons Between TL Methods}

\begin{table*}[!tp]
\setlength{\tabcolsep}{5.5pt}
\begin{minipage}[t]{.49\linewidth}
\centering
\medskip
 {\small
\setlength{\abovecaptionskip}{0.05cm}
\caption{\small The Result of Paraphrase Identification Task}
\label{tab:quoraresult}
\begin{tabular}{@{}llllll@{}}
\toprule
            & Prec@1 & Rec@1 & F1@1  & ACC     & AUC     \\ \midrule
Tgt-Only    & 0.717  & 0.551 & 0.623 & 0.792   & 0.822   \\
Src-Only    & 0.619  & 0.368 & 0.461 & 0.719   & 0.686   \\
Mixed       & 0.735  & 0.532 & 0.618 & 0.788   & 0.810   \\
Fine-Tune   & 0.713  & 0.567 & 0.632 & 0.790   & 0.825   \\
FS          & 0.734  & 0.595 & 0.657 & 0.797   & 0.831   \\
SS          & 0.744  & 0.601 & 0.665 & 0.800   & 0.837   \\
SS-Adv      & 0.743  & 0.603 & 0.666 & 0.808   & 0.842   \\\midrule
DRSS        & \textbf{0.757}  & 0.608 & 0.674$\dagger$ & \textbf{0.812}$\dagger$   & 0.847  \\
DRSS-Adv    & 0.753  & \textbf{0.620} & \textbf{0.680}$\dagger$ & 0.809  & \textbf{0.849}   \\\bottomrule
\end{tabular}}
\linespread{0.9}
\end{minipage}
\begin{minipage}[t]{0.5\linewidth}
\centering
\medskip
\small
\setlength{\abovecaptionskip}{0.05cm}
\caption{\small The Classification Result of NLI Task}
\label{tab:nliresult}
\begin{tabular}{@{}lllllll@{}}
\toprule
               & Fict. & Trav. & Gov. & Tele. & Slate & AVG   \\ \midrule
Tgt-Only       & 0.647   & 0.658  & 0.692      & 0.644     & 0.579 & 0.644 \\
Src-Only       & 0.520   & 0.516  & 0.540      & 0.520     & 0.488 & 0.517 \\
Mixed          & 0.647   & 0.647  & 0.675      & 0.648     & 0.580 & 0.639 \\
Fine-Tune      & 0.653   & 0.652  & 0.684      & 0.651     & 0.591 & 0.646 \\
FS            & 0.662   & 0.671  & 0.704      & 0.657     & 0.588 & 0.656 \\
SS            & 0.653   & 0.668  & 0.700      & 0.668     & 0.592 & 0.656 \\
SS-Adv        & 0.666   & 0.666  & 0.701      & 0.664     & 0.597 & 0.659 \\\midrule
DRSS          & 0.665   & \textbf{0.674}$\dagger$  & 0.706$\dagger$      & 0.673$\dagger$     & 0.605$\dagger$ & 0.665 \\
DRSS-Adv       & \textbf{0.676}$\dagger$   & 0.673  & \textbf{0.707}$\dagger$      & \textbf{0.675}$\dagger$     & \textbf{0.607}$\dagger$ & \textbf{0.668} \\  \bottomrule
\end{tabular}
\linespread{0.9}
\end{minipage}
\end{table*}

We further evaluated the performance of our transfer learning method in Table~\ref{tab:quoraresult} and Table~\ref{tab:nliresult}.

We can observe from Table~\ref{tab:nliresult} that for all the five target domains, Src-Only perform much worse than Tgt-Only, and the average performance of Mixed is even worse than Tgt-Only.
This implies that the source domain is quite different from all the target domains, and simply mixing the training data in two domains may lead to the model overfitting the source data since $|\mathcal{D}^\text{s}|$ is much larger than $|\mathcal{D}^\text{t}|$.
In addition, it is observed that the widely used Fine-Tune method can perform slightly better than Tgt-Only in most cases, which shows that pre-training the model parameters on a related source domain is better than randomly initializing them.
Moreover, in all the five domains, the performance of two existing transfer learning frameworks FS and SS are both 1.9\% better than that of Tgt-Only, which proves their usefulness.
Furthermore, our proposed method DRSS improves the average performance of SS to 0.665, and the improvements are significant over all the tasks with $p < 0.05$ based on McNemar's paired significance test.
This suggests that capturing the relationship between domains is generally useful for transfer learning.
Finally, we can see that the incorporation of adversarial loss into SS and DRSS further boosts their performance, and DRSS-Adv can achieve the best accuracy across all the methods.
Similar trends can be also observed for the PI task from Table~\ref{tab:quoraresult}.
Interestingly, by comparing Table~\ref{tab:indomain} and Table~\ref{tab:quoraresult}, we find that with the help of training data from the source domain, the performance of DRSS-Adv is even better than that of ESIM.
These observations demonstrate the effectiveness of our transfer learning method.

Apart from the effectiveness, we also measure the efficiency of each method. 
Since the first five methods only use a single hCNN model for prediction, the computational time is the same as hCNN.
As for SS, DRSS and their adversarial extensions,
the computational time is 6.9ms, which is slightly longer than the other five methods but still much shorter than LSTM-based methods as in Table~\ref{tab:indomain}.

\subsection{Domain Relationships}
\label{domrel}

\begin{figure}
\includegraphics[height=2in, width=3.2in]{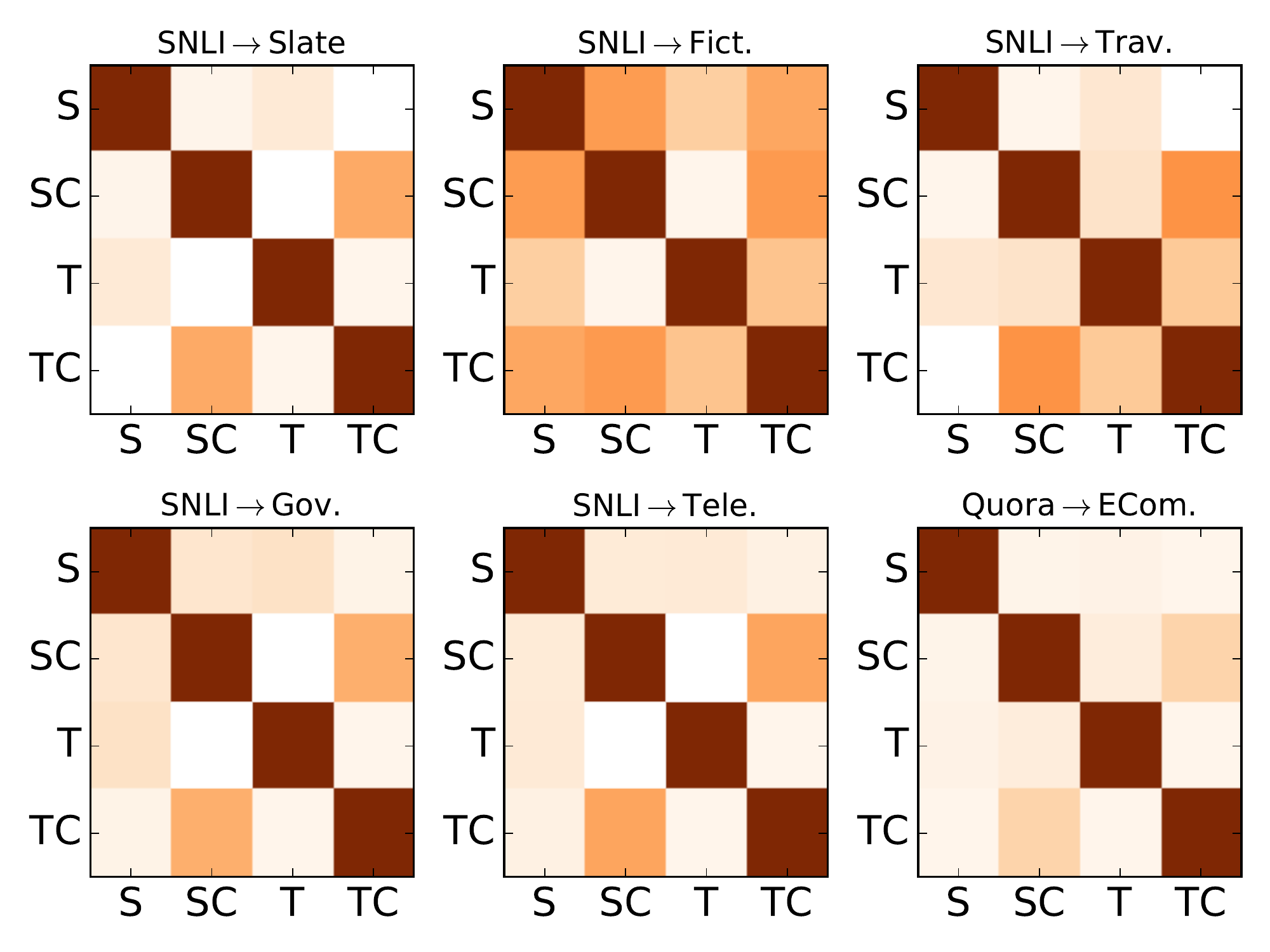}
\setlength{\abovecaptionskip}{-0.05cm}
\setlength{\belowcaptionskip}{-0.5cm}
\caption{Learnt Correlation Matrix. A darker color means a larger entry value. S:Source, T:Target, SC:Source-shared, TC: Target-shared.}
\label{fig:correlation}
\end{figure}
After obtaining the covariance matrix $\Omega$ for each source/target pair, we can derive their corresponding correlation matrices.
For better comparison, here we show the square root of the correlation matrices for DRSS.

As shown in Fig.~\ref{fig:correlation} that across all the six source/target pairs, $\mathbf{W_{sc}}$ and $\mathbf{W_{tc}}$ are positively related with each other. This is intuitive as the shared network is supposed to learn shared features between the source and the target domains, thus the learned $\mathbf{W_{sc}}$ and $\mathbf{W_{tc}}$ should be close to each other. This also shows the learned correlation matrix helps to capture the inter-domain relationship between $\mathbf{W_{sc}}$ and $\mathbf{W_{tc}}$.

In Fig.\ref{fig:correlation}, we can also see that for most source/target pairs except SNLI$\rightarrow$Fict, the correlation between $\mathbf{W_{s}}$ and $\mathbf{W_{sc}}$ and that between $\mathbf{W_{t}}$ and $\mathbf{W_{tc}}$ learnt by our model are with small values.
This indicates that in most cases, the shared feature space and the domain-specific feature space learnt by SS tend to be different from each other, and our model can help to reveal such intra-domain relationships. 

Finally, to help us get a deeper insight on the helpfulness of the adversarial training, we perform comparisons on the correlation matrices learnt by DRSS and DRSS-Adv.
We first show the result of SNLI$\rightarrow$Fict. in Table~\ref{tab:cor_score}.
As we can see, for DRSS, the correlation between $\mathbf{W_{s}}$ (or $\mathbf{W_{t}}$) and $\mathbf{W_{sc}}$ (or $\mathbf{W_{tc}}$) is relatively large,
while for DRSS-Adv, the correlation is relatively small.
For the other subtasks, we find that the learnt matrices of DRSS-Adv are similar to those of DRSS, but we still observe that the intra-domain correlations
of DRSS-Adv are generally smaller than those of DRSS.
This shows that adding the adversarial loss can encourage the shared feature space to capture more domain-independent features, and further make the shared and domain-specific feature spaces more different.
Therefore, the adversarial training can lead our model to better satisfy our assumption on the domain relationships, and finally improve the performance.
All the above observations demonstrate that our model can capture the \textit{inter-domain} and \textit{intra-domain} relationship as mentioned in Section~\ref{tl-model}.

\begin{table}[!tp]
\centering
\setlength{\abovecaptionskip}{0.05cm}
\caption{\small Correlation Matrices on SNLI$\rightarrow$Fict.}
\label{tab:cor_score}
\small
\begin{tabular}{@{}lcccccccc@{}}
\toprule
        & \multicolumn{4}{c}{DRSS} & \multicolumn{4}{c}{DRSS-Adv} \\
        \cmidrule(r){2-5} \cmidrule(r){6-9}
            & $\mathbf{W_{s}}$  & $\mathbf{W_{sc}}$  & $\mathbf{W_{t}}$   & $\mathbf{W_{tc}}$   & $\mathbf{W_{s}}$  & $\mathbf{W_{sc}}$  & $\mathbf{W_{t}}$   & $\mathbf{W_{tc}}$       \\ \midrule
$\mathbf{W_{s}}$    &1.000& 0.242& 0.101& 0.205   & 1.000& 0.090& 0.055& 0.062    \\
$\mathbf{W_{sc}}$   &0.242& 1.000& 0.008& 0.247   & 0.090& 1.000& 0.024& 0.221    \\
$\mathbf{W_{t}}$    &0.101& 0.008& 1.000& 0.127   & 0.055& 0.024& 1.000& 0.043     \\
$\mathbf{W_{tc}}$   &0.205& 0.247& 0.127& 1.000   & 0.062& 0.221& 0.043& 1.000    \\\bottomrule
\end{tabular}
\end{table}

\subsection{Extrinsic Evaluations}
\label{onlineeva}

As mentioned in Section~\ref{deploy}, for the online reranking algorithm, we propose to train GBDT by treating the prediction score of our DRSS model as another feature.
To achieve this, we first took out the prediction scores of our DRSS model on the validation set.
Then, we combined them together the other features as introduced in Section~\ref{deploy}, and trained GBDT on the validation set.
The model performance on the test set is reported in Table~\ref{tab:online}.
Note that the test time here denotes the average serving time (including the Response Time), which is different from the reported test time in Table~\ref{tab:indomain}.

As we can see from the offline test, the GBDT model with the feature derived from our DRSS model (referred to as GBDT-DRSS) is respectively 26.3\% and 7.1\% better than our existing online model (referred to as GBDT) and the GBDT model with the feature derived from hCNN (refered to as GBDT-hCNN) in $\text{F}_1$@1.
Although adding our DRSS feature leads to more computational time, the total prediction time is 80.7ms for each query question (i.e., QPS of 12), which is acceptable for our chatbots.

For online serving, to accelerate the computation, we set the number of candidates returned by \textit{Lucene} as 30, and bundle the 30 candidates into a mini-batch to feed into our model for prediction. 
For online evaluation, we randomly sampled 2750 questions, where 1317 questions are answered by GBDT and 1433 questions are answered by GBDT-DRSS.
Then, we asked one business analyst to annotate if the nearest question returned by models expresses the same meaning as the query question, and compared their precision at top-1.
As shown in Table~\ref{tab:online}, the Prec@1 of GBDT-DRSS is 18.8\% higher than that of GBDT.

\begin{table}[h!]
\centering
\setlength{\abovecaptionskip}{0.05cm}
\caption{\small The Performance of Online Serving }
\label{tab:online}
\footnotesize
\begin{tabular}{@{}lccc@{}}
\toprule
        & \multicolumn{2}{c}{Offline} & \multicolumn{1}{c}{Online Evaluation}\\
        \cmidrule(r){2-3} \cmidrule(r){4-4}
           &$\text{F}_1$@1  &Time(ms per query) & Prec@1 \\ \midrule
GBDT       &0.539 &20.1      & 0.614               \\
GBDT-hCNN  &0.636 &69.9      & -               \\
GBDT-DRSS  &\textbf{0.681} &80.7     &  \textbf{0.729}             \\\bottomrule
\end{tabular}
\end{table}

\section{Related Work}
\label{related}
\textbf{Paraphrase Identification and Natural Language Inference:}
Recent years have witnessed great successes of applying different neural networks, including Recursive Neural Networks (ReNN), Reccurrent Neural Networks (RNN) and Convolutional Neural Networks (CNN), into Paraphrase Identification and Natural Language Inference~\cite{socher:nips2011,snli:emnlp2015,yin:tacl2016}.
Although all these models have been shown to significantly outperform the traditional methods without deep learning, most of them only focus on improving the performance of standard in-domain setting, and therefore require a large amount of labeled data to train a robust model.
However, in practice, it will be time-consuming and costly to manually annotate much labeled data for each target domain we are interested in.
Hence, in this paper, our focus is to apply an efficient and effective NN model into a transfer learning framework so that we can leverage the large amount of labeled data from a related source domain to train a robust model for a resource-poor target domain, which can benefit our chatbot-based question answering system. 

\textbf{Transfer Learning:}
Transfer learning (TL) has been extensively studied in the last decade~\cite{long:tkde2014}.
Most existing studies for TL can be generally categorized into two groups.
The first line of work assumes that we have enough labeled data from a source domain and also a little labeled data from a target domain~\cite{daumeiii:acl2007}, and the second line assumes that we only have labeled data from source domain but may also have some unlabeled data from a target domain~\cite{blitzer:EMNLP2006,yu:EMNLP2016}.
Our study belongs to the first line of work, which is also referred to as \textit{supervised domain adaptation}.

For supervised domain adaptation, a majority of previous work belong to two clusters: instance-based and feature-based transfer learning.
The former focuses on mining from the source labeled data to find those instances that are similar to the distribution of the target domain, and combine them together with the target labeled data~\cite{dai:icml2007,jiang:acl2007}.
The core idea of the latter line of work is to find a shared feature space, which can reduce the divergence between the distribution of the source and the target domains~\cite{argyriou:nips2007,lee:icml2007,wang:icml2008,qiu:icdm2017}.
Our work follows the latter one, and tries to leverage NN models to learn a shared hidden representation for sentence pairs across domains.

\textbf{Deep Transfer Learning:} With the recent advances of deep learning, different NN-based TL frameworks have been proposed for image processing~\cite{yosinski:nips2014} and speech recognition~\cite{wang:apsipa2015} as well as NLP~\cite{mou:EMNLP2016,yang:iclr2017,liu:acl2017}.
A simple but widely used framework is referred to as \textit{fine-tuning} approaches, which first use the parameters of the well trained models on the source domain to initialize the model parameters of the target domain, and then fine tune the parameters based on labeled data in the target domain~\cite{yosinski:nips2014,mou:EMNLP2016}.
Another line of work can be referred to as \textit{multi-task feature learning} approaches, which bears the same intuition behind the feature-transfer methods as mentioned above.
Among this line of work, one typical framework is to simply use a shared NN to learn a shared feature space~\cite{mou:EMNLP2016,yang:iclr2017}, while another representative framework is to employ a shared NN and two domain-specific NNs to respectively derive a shared feature space and two domain-specific feature space~\cite{liu:acl2017}.
Motivated by the observation that both methods fail to consider the domain relationship, in this paper, we propose to jointly learn the shared feature representations and domain relationships in a unified model.
Moreover, inspired by the recent success of applying adversarial networks into unsupervised domain adaptation~\cite{ganin:jmlr2016,taigman:iclr2017} and multi-task learning~\cite{liu:acl2017}, we also incorporate the adversarial training into our transfer learning model in order to learn a more robust shared feature space across domains.

\section{Conclusions}
\label{conclusion}
In this paper, we systematically evaluated different base methods and transfer learning techniques for modelling sentence pairs, with the goal of proposing an effective and efficient transfer learning framework for PI and NLI. 
Specifically, we first proposed a hybrid CNN model on the basis of two existing models, and then further proposed a general transfer learning framework, which can simultaneously perform the shared feature learning and domain relationship learning in an end-to-end mode.
Evaluations on both a benchmark dataset and our own dataset 
showed that (1) our hybrid CNN model is both effective and efficient in comparison with several representative models; (2) our transfer learning framework can outperform all the existing frameworks across six source/target pairs.
We further deployed our transfer learning model in our online chatbot system, and showed that it can improve the performance of the existing system by a large margin.

\section*{Acknowledgements}

The authors would like to give great thanks to Feng Ji, Wei Zhou, Weipeng Zhao and Xu Hu
for their helpfulness during the project, and the anonymous reviewers for their constructive comments.

\bibliographystyle{ACM-Reference-Format}
\bibliography{sample}

\end{document}